%% file: main.tex
\documentclass[10pt,twocolumn,letterpaper]{article}

\usepackage{wacv}      


\definecolor{wacvblue}{rgb}{0.21,0.49,0.74}
\usepackage[pagebackref,breaklinks,colorlinks,allcolors=wacvblue]{hyperref}

\def\wacvPaperID{2663} 
\def\confName{WACV}
\def\confYear{2026}

\title{Large Sign Language Models: Toward 3D American Sign Language Translation}

\author{Sen Zhang$^{1}$\thanks{Equal contribution.} \quad
Xiaoxiao He$^{1}$\footnotemark[1] \quad Di Liu$^{2}$ \quad Zhaoyang Xia$^{1}$ \quad Mingyu Zhao$^{1}$ \quad Chaowei Tan$^{3}$ \\
\quad Vivian Li$^{6}$ \quad Bo Liu$^{4}$ \quad Dimitris N. Metaxas$^{2}$ \quad Mubbasir Kapadia$^{5}$ \vspace{+0.3em} \\
$^{1}$Rutgers University~~~$^{2}$Meta Reality Labs~~~$^{3}$Qualcomm~~~$^{4}$Walmart Global Tech \\$^{5}$Roblox~~~$^6$PRISMS\vspace{-0em} \\
}
\usepackage{stfloats}
\usepackage[utf8]{inputenc} 
\usepackage[T1]{fontenc}    
\usepackage{hyperref}       
\usepackage{url}            
\usepackage{booktabs}       
\usepackage{amsfonts}       
\usepackage{nicefrac}       
\usepackage{microtype}      
\usepackage{xcolor}         
\usepackage{graphicx}
\usepackage{tabularx}       
\usepackage{comment}
\usepackage{amsmath}
\usepackage{amssymb}
\usepackage{float}
\usepackage[symbol]{footmisc}
\usepackage{multirow}
\usepackage{xcolor}
\usepackage[]{hyperref}
\usepackage{listings}
\usepackage{xcolor} 
\usepackage{beramono} 
\usepackage{amsmath}
\lstdefinelanguage{json}{
    basicstyle=\ttfamily,
    numbers=left,
    numberstyle=\tiny, 
    stepnumber=1,
    numbersep=5pt,
    showstringspaces=false,
    breaklines=true,
    frame=lines,
    backgroundcolor=\color{gray!10},
    string=[s]"{\"},
    morestring=[b]",
    morecomment=[l]{//},
    morecomment=[s]{/*}{*/},
    commentstyle=\color{gray},
    keywordstyle=\color{blue},
    stringstyle=\color{red},
    identifierstyle=\color{black},
    keywords={true,false,null}
}

\begin{document}
\maketitle
\raggedbottom

\input{sec/0_abstract}    
\input{sec/1_intro}
\input{sec/2_relatedwork}
\input{sec/3_methods}
\input{sec/4_experiments}
\input{sec/5_discussion}

{
    \small
    \clearpage
    \bibliographystyle{ieeenat_fullname}
    \bibliography{main}
}

\clearpage
\appendix
\input{sec/6_appendix}

\end{document}

%% file: sec/0_abstract.tex
\begin{abstract}
We present Large Sign Language Models (LSLM), a novel framework for translating 3D American Sign Language (ASL) by leveraging Large Language Models (LLMs) as the backbone, which can benefit hearing-impaired individuals' virtual communication. Unlike existing sign language recognition methods that rely on 2D video, our approach directly utilizes 3D sign language data to capture rich spatial, gestural, and depth information in 3D scenes. This enables more accurate and resilient translation, enhancing digital communication accessibility for the hearing-impaired community. Beyond the task of ASL translation, our work explores the integration of complex, embodied multimodal languages into the processing capabilities of LLMs, moving beyond purely text-based inputs to broaden their understanding of human communication. We investigate both direct translation from 3D gesture features to text and an instruction-guided setting where translations can be modulated by external prompts, offering greater flexibility. This work provides a foundational step toward inclusive, multimodal intelligent systems capable of understanding diverse forms of language.
\end{abstract}

%% file: sec/1_intro.tex
\section{Introduction}
\label{sec:intro}

Sign languages are vital as they are the primary means of communication for hundreds of millions of hearing-impaired individuals globally~\citep{Davis2019HearingLoss}. Bridging the communication gap between sign language users and the non-signing majority is crucial for fostering inclusivity in education, employment, and daily social interactions. Recent advancements in deep learning and natural language processing have paved the way for developing automated sign language understanding and translation systems~\citep{Zhang2024DeepLearningSLR, Dinh2025SLRDataset,he2023dealing,liu2022transfusion,chang2022deeprecon,zhangli2022region,liu2021refined,liu2021label}, aiming to enhance accessibility and facilitate seamless communication. However, existing approaches relying on 2D video are inherently vulnerable to variations in camera view angles, lighting conditions, and complex backgrounds, which can degrade performance in 3D virtual environment scenarios and often necessitate constrained, simplified environments \citep{koller2020quantitative, cui2023spatial, Zheng2023CVTSLR, Kumari2024HybridSLR, Huang2024ResNetLSTM, Zuo2024OnlineCSLRT, Ranjbar2024IntraInterGloss, Hu2023SelfEmphasizingCSLR,he2024dice,Luo_2025_ICCV,liu2025lucas}. Concurrently, Large Language Models (LLMs) have demonstrated remarkable capabilities in understanding, generating, and translating textual languages. This success has spurred research into extending LLM capabilities to multimodal domains, including the interpretation of visual data. As highlighted by recent studies, efforts are underway to integrate LLMs for gloss-free sign language translation directly from video, aiming to leverage their extensive linguistic knowledge to produce more coherent and contextually appropriate translations \citep{gong2024llms, jimaging9110235, kim2024leveraging,xia2025visiar}.


\begin{figure*}[t]
\centering
\includegraphics[width=0.95\linewidth]{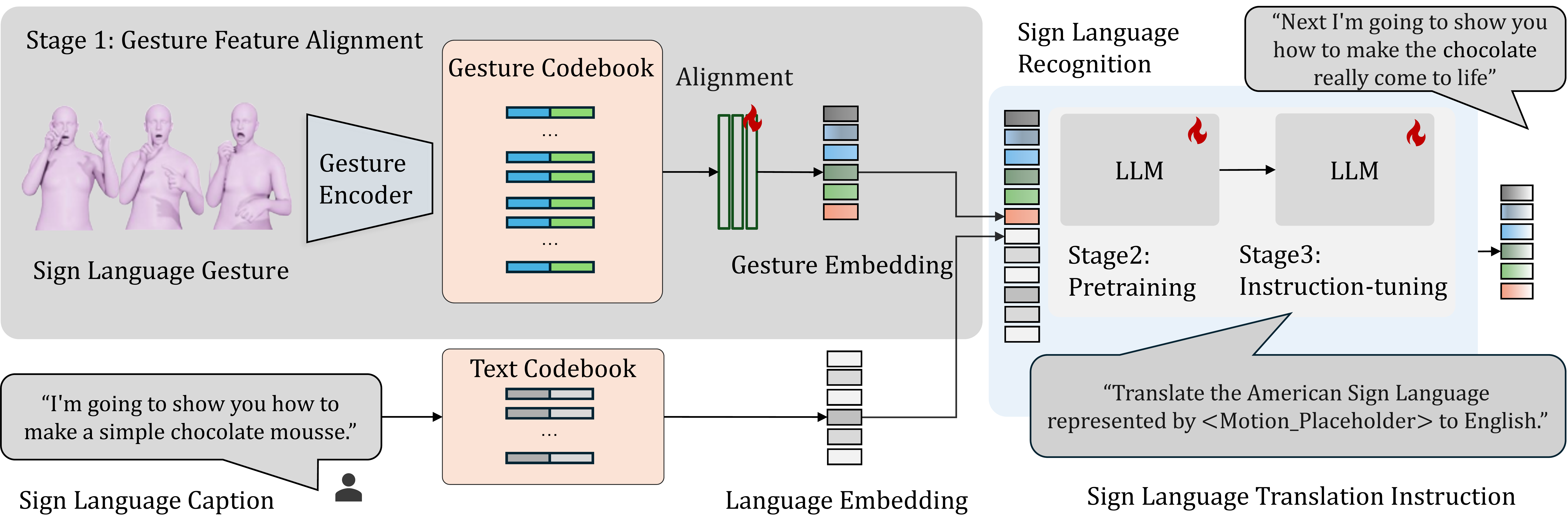}
\caption{The training pipeline for our Large Sign Language Model (LSLM) initiates with a pretrained text-based large language model and proceeds in three stages. First, we train a motion vector-quantized variational autoencoder (VQ-VAE); its quantized outputs are then used with a projection layer for gesture-text alignment. Second, these aligned gesture features serve as input to pretrain the LLM, enabling foundational sign language understanding. Finally, the model is instruction-finetuned to develop its ability to follow instructions with sign language.}
\label{fig:pipeline}
\end{figure*}

While prior work has predominantly relied on 2D video data for sign language translation, such approaches are constrained by fixed camera perspectives, simplified backgrounds, and limited spatial understanding~\cite{chen2025snapmoji,gao2025show,li2025hidden,dao2025improved,li2024implicit,zhangli2024resolving,zhangli2024layout,liu2024instantaneous,gao2024training,han2024proxedit,wen2024second,li2024steering,liu2023lepard,liu2023deformer,liu2023deep,martin2023deep}. To address these limitations for virtual communication, we propose a 3D-based approach that leverages rich motion representations encoded via SMPL-X~\citep{SMPL-X:2019}, as provided by recent 3D sign language datasets~\citep{yu2024signavatars}. These 3D representations offer detailed articulations of the body and hands, enabling more accurate modeling of the complex spatio-temporal patterns inherent in signing. Moreover, insights from LLM-driven human motion understanding and synthesis~\citep{jiang2024motiongpt, wang2024motiongpt} highlight the potential of combining temporal dynamics with large-scale language priors to interpret embodied communication. Ultimately, 3D sign language understanding not only enhances translation robustness but also enables hearing-impaired communication in virtual environment.

In this paper, we introduce the \textit{Large Sign Language Model (LSLM)}, a novel framework that harnesses the capabilities of large language models (LLMs) for translating 3D American Sign Language (ASL). Unlike traditional approaches that rely on 2D video data, our method leverages 3D gesture features derived from detailed body and hand motion. By interfacing these 3D motion representations with an LLM backbone, LSLM enhances translation robustness and accuracy for hearing-impaired users. More broadly, our work contributes to the emerging field of embodied multimodal understanding, expanding LLM capabilities beyond text to encompass complex, physical forms of human communication such as sign language.

Our contributions are threefold. \textbf{First}, we propose a novel framework for aligning 3D sign language representations with LLMs to enable end-to-end translation. \textbf{Second}, we explore direct translation from 3D gesture features to textual output, bypassing traditional gloss-based intermediates and reducing annotation dependency. \textbf{Third}, we introduce an instruction-guided translation setting that allows the model to adapt its outputs based on external prompts, offering greater flexibility and contextual awareness. Together, these advances represent a foundational step toward building inclusive, generalizable, and intelligent systems for sign language understanding and communication.

%% file: sec/2_relatedwork.tex
\section{Related Works}
\label{sec:relatedworks}

\subsection{Sign Language Recognition} 
Sign Language Recognition (SLR) has transitioned from sensor-based methods to robust, video-based deep learning frameworks for continuous signing, enabled by prevalent high-resolution video and affordable cameras. Computer vision techniques now extract crucial manual and non-manual features (e.g., hand shape, facial expressions)~\citep{bragg2019sign}. Early methods combining handcrafted features with Hidden Markov Models (HMMs) for spatio-temporal modeling were superseded by deep Convolutional Neural Networks (CNNs) for spatial feature extraction and Recurrent Neural Network (RNN)/Long Short-Term Memory (LSTM) models for temporal modeling~\citep{cui2017recurrent,koller2020quantitative}. Quantitative surveys document SLR's progression towards advanced deep learning architectures using full-frame video, iterative optimization, and data augmentation, enabling robust signer-independent recognition in unconstrained environments~\citep{koller2020quantitative,cui2023spatial}.

\subsection{Sign Language in Large Language Models} 
Recent video-based Sign Language Recognition (SLR) research integrates Large Language Models (LLMs) for gloss-free translation, bridging visual input and semantic language output~\citep{gong2024llms,jimaging9110235}. This paradigm employs transformer-based architectures to convert continuous sign videos directly into discrete token sequences, bypassing intermediate gloss annotations. Visual encoders compress spatio-temporal features into language-like representations~\citep{gong2024llms,jimaging9110235}. Innovations like the Vector-Quantized Visual Sign (VQ-Sign) module transform video inputs into character-level tokens via discrete codebook reconstruction and alignment, mitigating the visual-text modality gap and providing LLMs with structured input for leveraging their linguistic knowledge~\citep{kim2024leveraging,gong2024llms}. This gloss-free approach enables LLMs to resolve ambiguities and generate coherent translations from visual data, leading to scalable sign language translation systems~\citep{jimaging9110235,camgoz2020sign}. LLM integration signifies a paradigm shift, enabling seamless video-to-spoken language conversion by harnessing extensive textual understanding previously inaccessible to vision-only models~\citep{gong2024llms,camgoz2020sign}.

\subsection{LLM-based Motion Understanding} 
Recent advancements have seen the application of large language models (LLMs) to human motion synthesis. MotionGPT \citep{jiang2024motiongpt} treats the discrete tokens as context-free, raw indicators of motion which are then dynamically reinterpreted within subsequent generative modules. This design choice facilitates a more fluid and context-adaptive reconstruction process, one in which the generative network is free to reconstruct and refine motion details based solely on the current context and temporal requirements rather than being tethered to static, pre-learned representations. Such an approach is particularly beneficial in capturing the subtle nuances of signing—where minor variations in gesture intensity, duration, and transition phases can alter meaning—and serves to better accommodate inter-signer variability and environmental noise. Through such innovative synthesis, MotionGPT-type architectures \citep{jiang2024motiongpt,wang2024motiongpt} demonstrate significant promise in advancing the state-of-the-art in sign language recognition. Thus, we proposed Large Sign Language Model (LSLM), a novel framework specifically designed to adapt and extend these advanced motion-language principles to the unique challenges of this domain. Our LSLM focuses on interpreting the complex spatio-temporal dynamics and grammatical structures inherent in sign languages, by leveraging fine-grained motion representations inspired by these architectures and the contextual understanding capabilities of large language models, to achieve robust translation from sign inputs to spoken or written text.

%% file: sec/3_methods.tex
\section{Methods}
\label{sec:methods}

To enable sign language understanding, our method leverages the intrinsic linguistic knowledge of Large Language Models (LLMs). We address two primary technical challenges: 1) existing methods utilizing video as inputs are inefficient and cannot generalize to different camera angles or backgrounds; 2) the generation of sign language representations suitable for direct integration with LLM architectures. The subsequent subsections detail our proposed model architecture (Fig.~\ref{fig:pipeline}) and the training strategy designed to overcome these challenges.

\subsection{Environment Agnostic 3D Sign Language Tokenization}

Existing work, such as that by Gong et al.~\citep{gong2024llms}, often utilizes direct video feed as input for language models. While achieving commendable results on specific test datasets, these approaches can be susceptible to variations in visual artifacts or background changes, potentially leading to drastic fluctuations in the vision encoder's output. Furthermore, datasets used for training such models may exhibit strong consistency in camera angles, introducing a bias that can limit generalization. To address these challenges and enable sign language translation that is robust against domain shifts, we propose a sign language tokenizer centered on 3D skeletal data.

SMPL-X is a comprehensive, unified 3D model of the human body that distinctly represents the body, face, and hands together. SMPL-X features $10,475$ vertices and $54$ joints, which include articulations for the neck, jaw, eyeballs, and fingers, allowing for a richer representation of human motion and expression. However, in our sign language application, the two joints corresponding to the eyeballs are excluded, leading to the use of 52 joints for pose parameterization. The model is parameterized by pose, shape, and facial expression parameters, enabling fine-grained control over the generated 3D human mesh. This is especially  Assuming the availability of accurate 3D skeleton extraction from arbitrary video input, our core contribution is a vector quantized variational autoencoder (VQVAE)-based sign motion encoder. This encoder is designed to transform continuous sign language motion sequences into a discrete sequence of tokens. With existing SMPL-X parameter estimators \citep{cai2023smpler}, we are able to extract the human pose motion from a sign video and then utilize the motion representations for sign language understanding.

Our sign language tokenizer architecture, inspired by recent advancements in motion representation~\citep{jiang2024motiongpt}, consists of an encoder ($\mathcal{E}$) and a decoder ($\mathcal{D}$). The encoder ($\mathcal{E}$) takes a sequence of SMPL-X 3D skeleton representation, representing $M$ frames of human motion ($s^{1:M}$) to obtain latent vectors. These latent vectors ($\hat{z}^{1:L}$) are then subjected to a vector quantization process ($Q(\cdot)$). This involves a learnable codebook $C = \{c^k\}_{k=1}^{K}$, which contains $K$ latent embedding vectors, each of dimension $d$. Each latent vector $\hat{z}^i$ is mapped to its nearest codebook entry $c_k \in C$, resulting in a sequence of discrete motion tokens $z^{1:L}$, where $L$ is the downsampled length of the token sequence. 

\begin{align}
    z^{1:L} & = Q(\hat{z^{1:L}}) =  \text{min}_{c_k\in C} ||\hat{z}^i-c_k||_2
\end{align}

The decoder ($\mathcal{D}$) is then trained to reconstruct the original motion sequence $\hat{m}^{1:M}$ from these discrete tokens $z^{1:L}$. The training of the VQVAE involves a combination of reconstruction loss, an embedding loss to update the codebook vectors, and a commitment loss to ensure the encoder outputs commit to specific codebook entries \citep{guo2022tm2t,van2017neural}, defined in equation $(2)$, where $\text{sg}[\cdot]$ denotes the stop-gradient operation, and $\beta$ is a weighting factor.

\begin{figure*}[t]
    \centering
    \begin{gather}
    \mathcal{L}_{vq} = ||\hat{s}^{1:M}-s^{1:M}||_1+||\text{sg}[\mathcal{E}(s^{1:M})]-z^{1:L}||^2_2+\beta||\mathcal{E}(s^{1:M})-\text{sg}[z^{1:L}]||^2_2
    \end{gather}
\label{eqa:2}
\end{figure*}


\subsection{Sign Language Motion to Text Alignment}



Using this Sign Language tokenizer, we are able to encode the Sign Language to a sequence of quantized representations. However, for a pre-trained large language model (LLM) to understand and use sign language for applications, it needs to comprehend sign motion. The initial quantized embeddings from the tokenizer are not compatible with the word embeddings that the pre-trained LLM understands. Therefore, an alignment process is necessary. In our work, we utilize 2 multilayer perceptron (MLP) $\phi(\cdot)$ to align the quantized representation to the input embeddings of the LLM. After this mapping, these aligned embeddings $E_{sign}^{1:L}=\phi(z^{1:L})$ can be fused with the language embeddings $E_{lang}$ and feed into the Large Language Model, enabling it to process and understand the sign language input for tasks such as translation or recognition.

\begin{table*}[t]
\centering 
\caption{Quantitative results for sign language recognition during the pretraining stage.
MLP: \text{Direct Alignment and Fusion}; MLP+LLM: \text{Joint training}; MLP/LLM: \text{Staged training}.}
\renewcommand\tabcolsep{35pt} 
\resizebox{1\linewidth}{!}{
\begin{tabular}{@{}lcccc@{}}
\toprule
{Methods} & {Pretraining} & {Bleu@1$\uparrow$}&{Rouge$\uparrow$}&{CIDEr$\uparrow$} \\
\midrule
MotionGPT & - & $ 7.9 $ & $ 8.3 $ & $ 2.5 $ \\
Ours (Direct Alignment and Fusion) & MLP & $ 12.1 $ & $ 10.2 $ & $ 23.4 $ \\
Ours (Joint pretraining) & MLP+LLM & $ 13.6 $ & $ \textbf{11.7} $ & $ \textbf{26.9} $ \\
Ours (Staged pretraining) & MLP/LLM & $ \textbf{14.0} $ & $ \textbf{11.7} $ & $ 20.7 $ \\

\bottomrule
\end{tabular}%
}

\label{tab:pretrain-m2t-baseline}
\end{table*}

\subsection{Training Scheme}
\label{sec:training_scheme}
The training process is divided into three stages. The first stage is Sign Language Tokenizer training, which focuses on reconstructing the sign clips as discrete tokens.
The second stage is Modality-Alignment Pre-training, which aims to align the sign language space with pretrained text representation and facilitate collaboration across the SLR task. 
The third stage is Instruction Fine-Tuning, aimed at enhancing the model’s instruction-following capability.

\textbf{Sign Language Tokenizer Training.}
The initial phase involves the training of the Sign Language tokenizer. Upon successful training, any given sign language motion sequence, denoted as $s^{1:M}$, can be effectively represented as a corresponding sequence of motion tokens. Once this training stage is complete, the parameters of the motion tokenizer are fixed and remain unchanged throughout all subsequent stages of the processing pipeline.


\textbf{Pretraining.} The Llama models are trained and fine-tuned on natural language datasets. To enable these large language models to comprehend sign language, our methodology begins by first aligning motion embeddings from the sign language tokenizer with corresponding text embeddings through a dedicated training phase. Following this crucial alignment, we then continue to pre-train these models. This stage of pre-training utilizes the aligned sign language data and the corresponding text translation, with the specific aim of embedding sign language recognition capabilities within the large language models. For an input motion $s^{1:M}$, we consider the motion encoder trained from the previous section, which provides the motion feature $z^{1:L} = \mathcal{E}(s^{1:M})$. In this stage, Data was prepared in Gesture and Annotation pairs following the format presented in MotionGPT~\citep{jiang2024motiongpt}.


\textbf{Instruction Tuning.}
To enhance the translation of sign language into text, we developed an instruction-tuned dataset specifically for the sign language to text. This dataset was constructed by leveraging sign language corpora and reformulating the translation objective within an instruction-following paradigm, building upon foundational methodologies observed in text-motion alignment from broader motion capture research. Our focus is on the core task of converting sign language sequences into their corresponding textual representation. To this end, we composed dozens of varied instruction templates to promote robustness and prompt diversity. This process yielded a substantial collection of unique instructional prompts, each soliciting the spoken/written language equivalent of a given sign language sequence. Similar to the pretrain stage, we generate instructional ASL translation templates based on MotionGPT~\citep{jiang2024motiongpt}. Please refer to List~\ref{lst:merged_templates} in the appendix for examples.

%% file: sec/4_experiments.tex
\section{Experiments}

\subsection{Experimental Setup}
\label{sec:exp_setup}

\textbf{Datasets and Preprocessing.} We utilize the American Sign Language subset of SignAvatars~\citep{yu2024signavatars}, which provides 3D sign language motion sequences across multiple languages with corresponding text annotations, and statistical information is shown in Tab~\ref{tab:asl-dataset} in the appendix. How2Sign dataset contains over 30000 samples, and we divided the dataset into 80\% training set, 10\% test, and validation set. The 3D motion is represented via the SMPL-X model, using 52 joints for body and hand kinematics. Following the HumanML3D~\citep{t2m} processing methodology, we extract a 623-dimensional feature vector per frame. This vector encodes root velocity, local joint positions, and local joint rotations, serving as the input representation for our model.

\begin{figure}[b]
\centering
\vspace{-10pt}
\includegraphics[width=\linewidth]{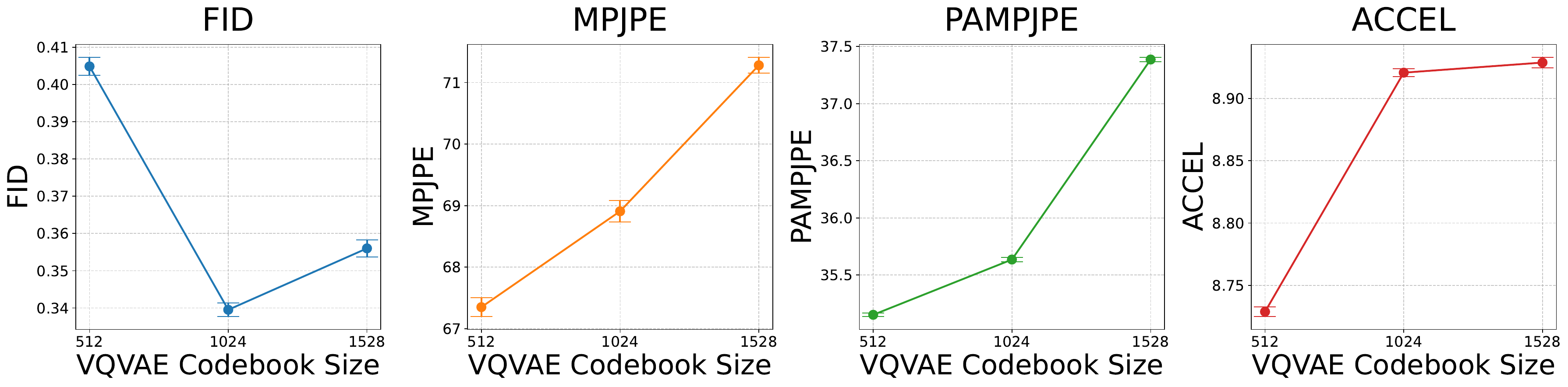}
\vspace{-10pt}
\caption{Evaluation of VQ-VAE codebook sizes on the How2Sign 3D dataset. A codebook size of 1024 offers the best trade-off, achieving the lowest FID while maintaining competitive MPJPE and PAMPJPE scores.}
\label{fig:vqvae}
\end{figure}

\textbf{Evaluation Metrics.}  We evaluate translation quality using metrics standard in sign language and motion-to-text research, following practices in works~\citep{cui2017recurrent,koller2020quantitative,cui2023spatial,gong2024llms,jimaging9110235,camgoz2020sign,kim2024leveraging} and MotionGPT~\citep{jiang2024motiongpt}. We report Word Error Rate (WER) to measure lexical accuracy against reference translations (lower is better). Additionally, we utilize the established linguistic metrics BLEU \citep{papineni-etal-2002-bleu}, ROUGE-L~\citep{lin-2004-rouge}, and CIDEr~\citep{Vedantam2015CIDEr} to assess translation fluency, content overlap, and semantic similarity, respectively.

\textbf{Implementation Details.} For gesture tokenization, we set the codebook size to 1024. This value was empirically determined to provide the optimal balance for reconstruction quality, achieving the lowest Fréchet Inception Distance (FID) while maintaining favorable Mean Per Joint Position Error (MPJPE) and Percentage of Accelerated MPJPE (PAMPJPE), with detailed metric comparisons presented in Fig.~\ref{fig:vqvae}. The gesture tokenizer was trained using the AdamW optimizer~\citep{adamw} with a learning rate of $2 \times 10 ^{-4}$ and a batch size of 256. In the subsequent gesture-language alignment stage, \texttt{Llama-3.2-3B-Instruct}~\citep{MetaLlama3.2Blog2024} was employed as the large language model backbone, and we set learning rate to $1 \times 10 ^{-6}$ for modality alignment, and $5 \times 10 ^{-7}$ for LLM backbone fine-tuning. Our experiments are conducted on 8 Nvidia A100 GPUs.

\begin{table*}[!htbp]
\centering 
\caption{Quantitative results for sign language recognition during the instruction-tuning stage. MLP: \text{Direct Alignment and Fusion}; MLP+LLM: \text{Joint training}.}
\renewcommand\tabcolsep{30pt} 
\resizebox{1\linewidth}{!}{
\begin{tabular}{@{}lcccccc@{}}
\toprule
\multirow{1}{*}{Methods}& {Pretrained Modules}& {Instruction Tuning}& {Bleu@1$\uparrow$}&{Rouge$\uparrow$}&{CIDEr$\uparrow$} \\
\midrule 

MotionGPT & - & - &
$ 2.5 $ & $ 2.4 $& $ 5.7 \times e^{-4} $ \\
Ours & MLP & LLM &
$ \textbf{14.2} $ & $ \textbf{11.9} $& $ \textbf{28.0} $ \\
Ours & MLP & MLP+LLM &
$ 13.8 $ & $ 11.7 $& $ 21.0 $ \\
Ours & MLP+LLM & LLM &
$ 14.1 $ & $ 11.5 $& $ 20.6 $  \\
Ours & MLP+LLM & MLP+LLM &
$ 14.1 $ & $ 11.2 $& $ 27.3 $ \\

\bottomrule
\end{tabular}%
}

\label{tab:slr-instruction-baseline}
\end{table*}






\subsection{Comparison with Motion-to-Text Approaches}

To establish a baseline for Sign Language Recognition (SLR) task, we adapted the Motion-to-Text (M2T) capabilities of the MotionGPT framework~\citep{jiang2024motiongpt}, training it on 3D American Sign Language (ASL) data. This adaptation is motivated by the inherent structural similarity between M2T and SLR, as both tasks involve translating sequences of motion or gesture into textual output. However, a crucial distinction lies in the semantic nature of the target text: M2T typically aims to generate descriptive captions of the observed motion, whereas SLR endeavors to produce direct and accurate translations of the signed gestures. 

As shown in Table~\ref{tab:pretrain-m2t-baseline}, our proposed Large Sign Language Model (LSLM) consistently outperforms the adapted MotionGPT baseline across all pretraining metrics.  This superior performance underscores LSLM's enhanced capacity for comprehending the complex nuances of ASL gesture modality, which consequently leads to the generation of more accurate translations. Two key architectural differences contribute to this performance gain. \textit{1) Enhanced Motion Representation}: Unlike MotionGPT, which directly inputs discrete motion tokens into its LLM backbone, LSLM utilizes gesture embeddings derived after the initial tokenization process. \textit{2) Explicit Modality Alignment}: LSLM incorporates a dedicated modality alignment mechanism, \textit{i.e.}, Direct Alignment and Fusion. This component is designed to map the learned embeddings from the gesture space to the text embedding space of the Llama LLM backbone, facilitating a more effective and coherent integration of visual-gestural information with the language model's existing linguistic knowledge.

We further compare the models under instruction-following conditions, with results summarized in Table~\ref{tab:slr-instruction-baseline}. Following the initial pertaining, LSLM demonstrated effective instruction fine-tuning, evidenced by a consistent improvement in performance metrics compared to its pre-instruction fine-tuning state. In stark contrast, the MotionGPT baseline exhibited a degradation in performance after its instruction fine-tuning phase. This divergence indicates the superior adaptability and effectiveness of our LSLM framework in leveraging instructional cues for the nuanced task of sign language translation, further validating our architectural and training choices.

\begin{table*}[t]
\centering
\caption{Ablation studies of the pretrain scheme for the Sign Language Recognition.
MLP: \text{Direct Alignment and Fusion}; MLP+LLM: \text{Joint training}; MLP/LLM: \text{Staged training}.
}

\renewcommand\tabcolsep{22pt} 
\resizebox{1\linewidth}{!}{
\begin{tabular}{@{}lcccccccc@{}} 
\toprule
\multirow{1}{*}{Pretraining}&
{Bleu@1$\uparrow$}&{Bleu@4$\uparrow$}&{Rouge$\uparrow$}&{CIDEr$\uparrow$}&{WER$\downarrow$}&{Insertions$\downarrow$}&{Deletions$\downarrow$} \\
\midrule

MLP&
$ 12.1 $ & $ 1.0 $& $ 10.2 $ & $ 23.4 $ & $ 140 $ & $ 4.2 $ & $ 4.0 $ \\
MLP+LLM &
$ 13.6 $ & $ 1.5 $& $ \textbf{11.7} $ & $ \textbf{26.9} $ & $ \textbf{114} $  & $ 4.0 $ & $ 1.7 $\\
MLP/LLM &
$ \textbf{14.0} $ & $ \textbf{1.6} $& $ \textbf{11.7} $ & $ 20.7 $ & $ 118 $ & $ \textbf{3.9} $ & $ \textbf{1.4} $ \\

\bottomrule
\end{tabular}%
}
\label{tab:SLR}
\end{table*}

\subsection{Modality Alignment and Pretraining} 

To determine the most effective training scheme for aligning the gesture modality with the Large Language Model (LLM) backbone in the end-to-end Sign Language Recognition (SLR) task, we conducted three ablation studies. For these experiments, the data were structured as gesture-translation pairs, prepared according to the procedures detailed in Section~\ref{sec:training_scheme}, which is also happens to be the format of Direct Sign Language Recognition. The following three training schemes were systematically investigated: \textbf{1) Direct Alignment and Fusion}: This approach involved a straightforward alignment of the gesture feature space with the text embedding space using a Multi-Layer Perceptron (MLP). The resulting gesture embeddings were then fused with the LLM's input embeddings for processing by the LLM backbone. \textbf{2) Joint pretraining }: In this scheme, both the MLP (for gesture-to-text space alignment) and the entire LLM backbone were jointly pretrained. This was performed using the paired gesture and textual annotation data, allowing for simultaneous optimization of both the gesture processing component and the language model. \textbf{3) Staged pretraining}: This method adopted a two-stage process. Initially, the MLP was pretrained to learn the mapping from gesture features to an intermediate representation aligned with the LLM's embedding space. Subsequently, during the second stage, the LLM backbone was pretrained on the translation task while the parameters of the pretrained MLP were kept frozen.

The outcomes of our ablation studies, detailed in Table~\ref{tab:SLR}, provide insights into the efficacy of different training schemes. The Direct Alignment and Fusion approach yielded decent results over the baseline. This initial success indicates the importance and effectiveness of explicitly aligning the gesture modality with the LLM's textual embedding space. But compared to the latter two schemes, this scheme generates redundant words as the deletion is higher. Building upon this observation, the Joint Petraining scheme demonstrated further improvements in textual similarity metrics, notably BLEU, ROUGE, and WER. This suggests that allowing the LLM to adapt concurrently with the gesture modality facilitates a more refined translation output. Conversely, the Staged Pretraining approach, while conceptually sound, resulted in a discernible decrease in semantic similarity scores (e.g., CIDEr), indicating that freezing the MLP after pretraining may restrict the LLM's ability to fully leverage the learned gesture representations during its own fine-tuning phase. Considering the overall balance of performance across all metrics, we selected the Joint Pretraining strategy as the optimal configuration for LSLM on the SLR task. Approach 1 and 3 checkpoints were subsequently employed for further instructional experiments and evaluations. 

\begin{table*}[t]
\centering
\caption{Ablation studies of the instruction-tuning scheme for the Sign Language Recognition. MLP: \text{Direct Alignment and Fusion}; MLP+LLM: \text{Joint training}.}
\renewcommand\tabcolsep{15pt} 
\resizebox{1\linewidth}{!}{
\begin{tabular}{@{}lcccccccc@{}} 
\toprule
\multirow{1}{*}{Pretraining}&{Instruction Tuning}&
{Bleu@1$\uparrow$}&{Bleu@4$\uparrow$}&{Rouge$\uparrow$}&{CIDEr$\uparrow$}&{WER$\downarrow$}&{Insertions$\downarrow$}&{Deletions$\downarrow$} \\
\midrule

MLP & LLM &
$ \textbf{14.2} $ & $ \textbf{1.6} $& $ \textbf{11.9} $ & $ \textbf{28.0} $ & $ 146 $ & $ 4.0 $ & $ 1.0 $ \\
MLP & MLP+LLM &
$ 13.8 $ & $ 1.5 $& $ 11.7 $ & $ 21.0 $ & $ 149 $ & $ 4.4 $ & $ \textbf{0.9} $ \\
MLP+LLM & LLM &
$ 14.1 $ & $ 1.5 $& $ 11.5 $ & $ 20.6 $ & $ 141 $ & $ 3.4 $ & $ 1.2 $ \\
MLP+LLM & MLP+LLM &
$ 14.1 $ & $ 1.5 $& $ 11.2 $ & $ 27.3 $ & $ \textbf{132} $ & $ \textbf{3.0} $ & $ 1.4 $ \\

\bottomrule
\end{tabular}%
}
\label{tab:slr-instruction}
\end{table*}

\begin{figure*}[!htbp]
\centering
\includegraphics[width=0.85\linewidth]{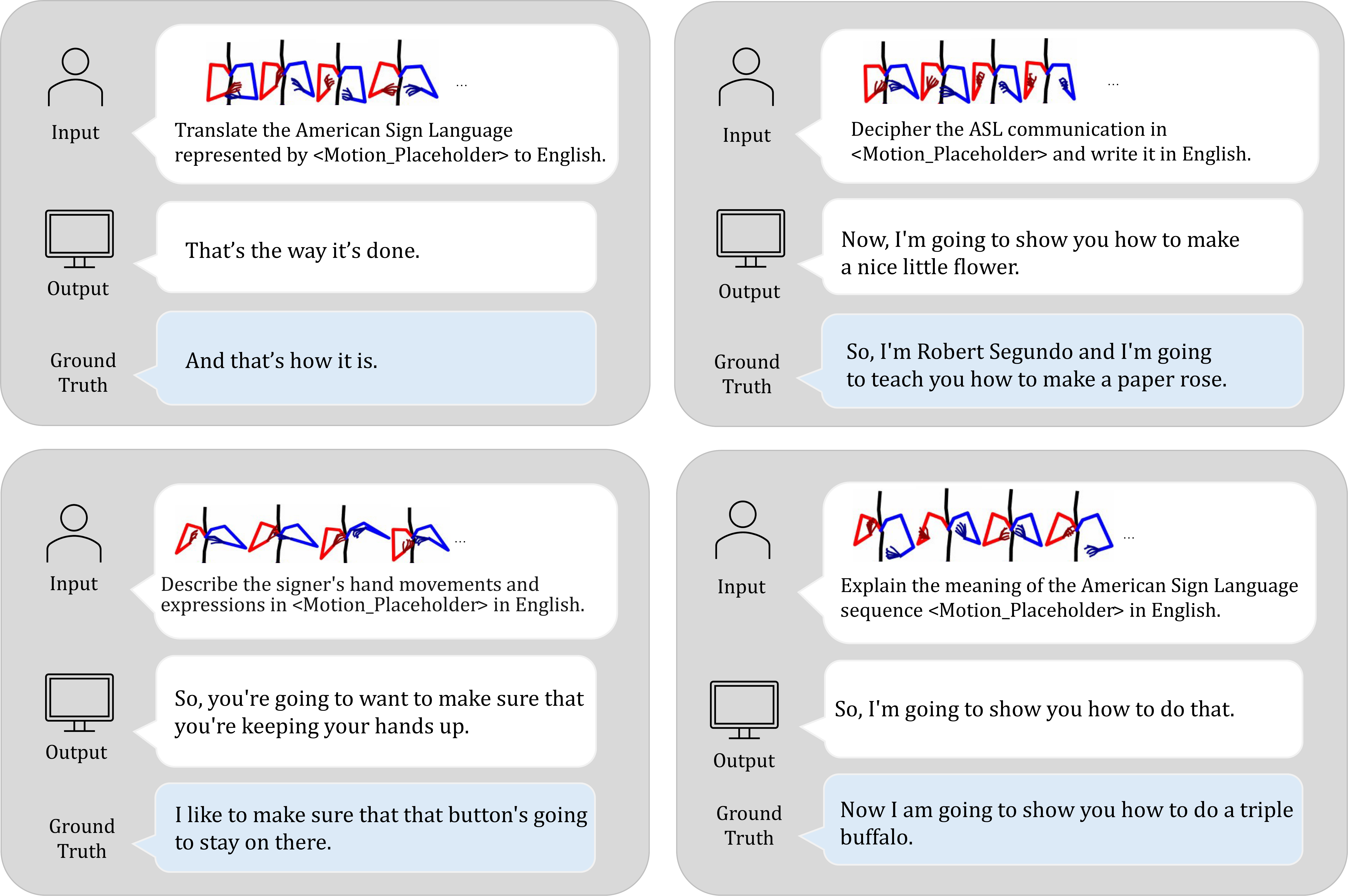}
\caption{Instruction-guided Sign Language Recognition (SLR) examples from our LSLM framework. Each case shows the model's translation given a gesture sequence and prompt, alongside the ground truth. Outputs may slightly differ in wording but preserve the core meaning.}

\label{fig:vis_slr}
\end{figure*}

\subsection{Instruction Finetuning}

To further enhance the versatility and applicability of LSLM beyond direct translation, we incorporated an instruction fine-tuning stage. This stage aims to enable the model to perform the sign language recognition task in response to varied explicit instructions, thereby broadening its utility for more diverse interactive scenarios. For this purpose, we adapted instruction templates originally proposed in MotionGPT~\citep{jiang2024motiongpt}. These templates, initially designed for general, descriptive motion, were systematically transformed into formats suitable for ASL translation tasks, as detailed in Section~\ref{sec:training_scheme}. To identify the optimal approach for this stage, we also conducted a series of ablation studies focused on different fine-tuning schemes specifically for instruction following.

Further analysis of instruction fine-tuning strategies, with results detailed in Table~\ref{tab:slr-instruction}, revealed interesting trade-offs. Notably, the Direct Alignment and Fusion scheme, when followed by an LLM-only fine-tuning stage dedicated to instruction following, achieved the highest performance on BLEU, ROUGE, and CIDEr metrics, despite not being the top performer in the initial pretraining phase. The second-best performance was observed with the Joint Pretraining approach when augmented with an additional joint instruction tuning phase. Other hybrid or 'mix-and-match' fine-tuning strategies that were explored did not surpass the efficacy of these two leading methods. Consequently, based on these comprehensive results, our LSLM framework adopts the Direct Alignment and Fusion approach for initial gesture modality alignment, followed by an LLM-only instruction fine-tuning stage for the sign language recognition with instructions task. Illustrative translation samples from this configuration are presented in Fig.~\ref{fig:vis_slr}.

\section{Ablation Study on LLM Backbone} 

We conduct several ablation studies to evaluate the effectiveness of \texttt{Qwen2.5-3B-Instruct}~\cite{qwen25technicalreport} as the LLM backbone for LSLM. We trained our model with the same learning rates and setup as mentioned in Sec~\ref{sec:exp_setup}.


\subsection{Pretraining}

In the pretraining stage, we analyzed a Qwen-based backbone against the established baseline, with results detailed in Table~\ref{tab:pretrain-m2t-baseline-qwen}. Similar to observations with the Llama backbone, employing Qwen led to notable improvements in fluency and content coverage, as indicated by higher BLEU and ROUGE scores. Furthermore, there was a clear enhancement in semantic similarity (e.g., CIDEr). However, when directly compared, the Qwen backbone did not surpass the performance levels achieved by the Llama backbone across the evaluated metrics, as shown in Table~\ref{tab:slr-qwen}.

\begin{table*}[htbp]
\centering 
\caption{Quantitative results for Qwen as LLM backbone for SLR during the pretraining stage.
MLP: \text{Direct Alignment and Fusion}; MLP+LLM: \text{Joint training}; MLP/LLM: \text{Staged training}.}
\renewcommand\tabcolsep{35pt} 
\resizebox{1\linewidth}{!}{
\begin{tabular}{@{}lcccc@{}}
\toprule
{Methods} & {Pretraining} & {Bleu@1$\uparrow$}&{Rouge$\uparrow$}&{CIDEr$\uparrow$} \\
\midrule
MotionGPT & - & $ 7.9 $ & $ 8.3 $ & $ 2.5 $ \\
Qwen (Direct Alignment and Fusion) & MLP & $ 8.2 $ & $ 8.4 $ & $ 12.5 $ \\
Qwen (Joint pretraining) & MLP+LLM & $ \textbf{8.5} $ & $ \textbf{9.8} $ & $ \textbf{18.5} $ \\

\bottomrule
\end{tabular}%
}

\label{tab:pretrain-m2t-baseline-qwen}
\end{table*}

Regarding lexical accuracy, the Qwen backbone demonstrated a higher Word Error Rate (WER) compared to Llama. Further analysis of the generation patterns indicated that Qwen tended to produce more non-relevant words, resulting in a significantly higher insertion rate than observed with the Llama-based model.

\begin{table*}[htbp]
\centering
\caption{Comparison of LLM backbones for the Sign Language Recognition.
MLP: \text{Direct Alignment and Fusion}; MLP+LLM: \text{Joint training}; MLP/LLM: \text{Staged training}.
}

\renewcommand\tabcolsep{20pt} 
\resizebox{1\linewidth}{!}{
\begin{tabular}{@{}lcccccccc@{}} 
\toprule
\multirow{1}{*}{Pretraining}&
{Bleu@1$\uparrow$}&{Bleu@4$\uparrow$}&{Rouge$\uparrow$}&{CIDEr$\uparrow$}&{WER$\downarrow$}&{Insertions$\downarrow$}&{Deletions$\downarrow$} \\
\midrule

MLP (Llama)&
$ 12.1 $ & $ 1.0 $& $ 10.2 $ & $ 23.4 $ & $ 140 $ & $ 4.2 $ & $ 4.0 $ \\
MLP+LLM (Llama)&
$ 13.6 $ & $ 1.5 $& $ \textbf{11.7} $ & $ \textbf{26.9} $ & $ \textbf{114} $  & $ 4.0 $ & $ 1.7 $\\
MLP/LLM (Llama)&
$ \textbf{14.0} $ & $ \textbf{1.6} $& $ \textbf{11.7} $ & $ 20.7 $ & $ 118 $ & $ \textbf{3.9} $ & $ \textbf{1.4} $ \\

MLP (Qwen)&
$ 8.2 $ & $ 0.3 $& $ 8.3 $ & $ 12.5 $ & $ 315 $ & $ 9.1 $ & $ 3.5 $ \\
MLP+LLM (Qwen)&
$ 8.5 $ & $ 0.8 $& $ 9.8 $ & $ 18.5 $ & $ 185 $  & $ 10.0 $ & $ 1.9 $\\

\bottomrule
\end{tabular}%
}
\label{tab:slr-qwen}
\end{table*}

\subsection{Instruction Tuning}

In the instruction tuning phase, the Qwen backbone also demonstrated improved performance compared to the baseline, following similar trends observed with Llama as shown in Table~\ref{tab:slr-instruction-baseline-qwen}. Specifically, when using Qwen, we achieved strong semantic similarity scores with the Direct Alignment and Fusion strategy for initial modality mapping, followed by LLM-only fine-tuning for instruction tuning. Alternatively, for better lexical accuracy, joint training during both the initial alignment and instruction fine-tuning stages produced favorable outcomes with the Qwen backbone.

\begin{table*}[htbp]
\centering 
\caption{Quantitative results for sign language recognition with instruction for Qwen as backbone. MLP: \text{Direct Alignment and Fusion}; MLP+LLM: \text{Joint training}.}
\renewcommand\tabcolsep{30pt} 
\resizebox{1\linewidth}{!}{
\begin{tabular}{@{}lcccccc@{}}
\toprule
\multirow{1}{*}{Methods}& {Pretrained Modules}& {Instruction Tuning}& {Bleu@1$\uparrow$}&{Rouge$\uparrow$}&{CIDEr$\uparrow$} \\
\midrule 

MotionGPT & - & - &
$ 2.5 $ & $ 2.4 $& $ 5.7 \times e^{-4} $ \\
Ours(Qwen) & MLP & LLM &
$ 10.2 $ & $ 11.2 $& $ \textbf{22.3} $ \\
Ours(Qwen) & MLP & MLP+LLM &
$ 10.0 $ & $ 11.1 $& $ 18.6 $ \\
Ours(Qwen) & MLP+LLM & LLM &
$ 12.9 $ & $ 11.5 $& $ 15.1 $  \\
Ours(Qwen) & MLP+LLM & MLP+LLM &
$ \textbf{13.2} $ & $ \textbf{11.8} $& $ 15.1 $ \\
\bottomrule
\end{tabular}%
}
\label{tab:slr-instruction-baseline-qwen}
\end{table*}

Consistent with patterns observed with the Llama backbone in Table~\ref{tab:slr-instruction-qwen}, the Qwen-based LSLM achieved its highest semantic similarity scores using the Direct Alignment and Fusion strategy followed by LLM-only fine-tuning. Similar to its pretraining performance, the Qwen-based model exhibited higher insertion rates. While this suggests a tendency toward generating more verbose translations, the model still maintained comparable lexical accuracy metrics in certain configurations.

\begin{table*}[b]
\centering
\caption{LLM Backbone comparison for SLR with instruction. MLP: \text{Direct Alignment and Fusion}; MLP+LLM: \text{Joint training}.}
\renewcommand\tabcolsep{10pt} 
\resizebox{1\linewidth}{!}{
\begin{tabular}{@{}lcccccccccc@{}} 
\toprule
\multirow{1}{*}{Backbone}&{Pretraining}&{Instruction Tuning}&
{Bleu@1$\uparrow$}&{Bleu@4$\uparrow$}&{Rouge$\uparrow$}&{CIDEr$\uparrow$}&{WER$\downarrow$}&{Insertions$\downarrow$}&{Deletions$\downarrow$} \\
\midrule

Llama & MLP & LLM &
$ \textbf{14.2} $ & $ 1.6 $& $ \textbf{11.9} $ & $ \textbf{28.0} $ & $ 146 $ & $ 4.0 $ & $ 1.0 $ \\
Llama & MLP & MLP+LLM &
$ 13.8 $ & $ 1.5 $& $ 11.7 $ & $ 21.0 $ & $ 149 $ & $ 4.4 $ & $ 0.9 $ \\
Llama & MLP+LLM & LLM &
$ 14.1 $ & $ 1.5 $& $ 11.5 $ & $ 20.6 $ & $ 141 $ & $ 3.4 $ & $ 1.2 $ \\
Llama & MLP+LLM & MLP+LLM &
$ 14.1 $ & $ 1.5 $& $ 11.2 $ & $ 27.3 $ & $ \textbf{132} $ & $ \textbf{3.0} $ & $ 1.4 $ \\
Qwen & MLP & LLM &
$ 10.2 $ & $ 1.2 $& $ 11.2 $ & $ 22.3 $ & $ 195 $ & $ 10.1 $ & $ 1.0 $ \\
Qwen & MLP & MLP+LLM &
$ 10.0 $ & $ 1.2 $& $ 11.1 $ & $ 18.6 $ & $ 188 $ & $ 9.9 $ & $ 1.1 $ \\
Qwen & MLP+LLM & LLM &
$ 12.9 $ & $ \textbf{1.9} $& $ 11.5 $ & $ 15.1 $ & $ 149 $ & $ 5.1 $ & $ \textbf{0.8} $ \\
Qwen & MLP+LLM & MLP+LLM &
$ 13.2 $ & $ \textbf{1.9} $& $ 11.8 $ & $ 15.1 $ & $ 150 $ & $ 5.1 $ & $ \textbf{0.8} $ \\

\bottomrule
\end{tabular}%
}
\label{tab:slr-instruction-qwen}
\end{table*}

%% file: sec/5_discussion.tex
\section{Discussion}

In this paper, we introduce the Large Sign Language Model (LSLM), a novel framework that successfully addresses 3D Sign Language translation for virtual communication.
We solve this by pioneering the alignment of robust 3D sign gesture features, derived from the SMPL-X representations in 3D ASL datasets, with a Large Language Model backbone. This integration enables LSLM to achieve end-to-end translation of 3D American Sign Language, make initial attempts in direct gesture-to-text translation without traditional gloss intermediaries, and explore instruction-guided translation for greater output control, thereby enhancing translation accuracy for hearing-impaired individuals and advancing LLMs' capacity to understand complex, embodied multimodal communication.

\textbf{Limitations and Broader Impacts.} Large Language Models (LLMs) face significant challenges and limitations when attempting to effectively understand sign language. Unlike spoken or written languages that LLMs excel at, sign languages are inherently visual-spatial, relying on handshapes, movements, facial expressions, and body posture to convey meaning. This multi-modal nature presents a substantial hurdle for current LLM architectures, which are primarily text-based. Another obstacle is the fundamental difference in data representation and the need for vision and pattern recognition capabilities to accurately interpret and translate the nuances of sign. Furthermore, unlike general motion datasets~\citep{AMASS:ICCV:2019,Plappert2016kit}, the relative scarcity of large, high-quality, and diverse sign language datasets, compared to text or general motion data, hinders the training of robust LLMs. Overcoming these borders would ultimately pave the way for more inclusive and accessible technology for Deaf and hard-of-hearing communities. Future work could focus on expanding diverse 3D ASL datasets and developing linguistically-aware augmentation and learning strategies for models like LSLM. 

%% file: sec/6_appendix.tex
\begin{figure*}[t]
    \begin{center}
        {\Large\bfseries Supplementary Material for Large Sign Language Models: Toward 3D American Sign Language Translation\par}
    \end{center}
    \vspace{-0.5cm} 
\end{figure*}

\section{Implementation details}

\subsection{Training Hyperparameters}
\label{app:training_hyperparameters}
Key hyperparameters for our model training are as follows:

\begin{table}[h]
\centering
\caption{Training Hyperparameters}
\renewcommand\tabcolsep{25pt} 
\resizebox{1\linewidth}{!}{
\begin{tabular}{lll}
\toprule
\textbf{Component} & \textbf{Hyperparameter} & \textbf{Value/Type} \\
\midrule
\multirow{5}{*}{\textbf{Gesture Tokenizer (VQ-VAE)}} 
& Codebook Size & 1024 \\
& Optimizer & AdamW \\
& Learning Rate & $2 \times 10^{-4}$ \\
& Batch Size & 256 \\
& Number of Joints & 52 \\
\midrule
\multirow{4}{*}{\textbf{LLM + Alignment}} 
& Max Token Length & 250 \\
& Batch Size per GPU & 16 \\
& Alignment MLP LR & $1 \times 10^{-6}$ \\
& LLM Fine-tune LR & $5 \times 10^{-7}$ \\
\midrule
\multirow{2}{*}{\textbf{Motion Token IDs}} 
& LLaMA & \texttt{128259} \\
& Qwen & \texttt{151668} \\
\bottomrule
\end{tabular}}
\end{table}

\subsection{Dataset Detail}
\label{app:training_hyperparameters}
The details of the SignAvatar dataset are listed below:

\begin{table}[h]
\centering 
\caption{Number of samples in SignAvatars dataset.}
\label{tab:asl-dataset}
\renewcommand\tabcolsep{25pt} 
\resizebox{1\linewidth}{!}{
\begin{tabular}{@{}lcccc@{}} 
\toprule
Dataset samples & {Dev} & {Test} & {Val} & {Language} \\ 
\midrule
How2Sign~\citep{how2sign}  & $ 24476 $ & $ 3060 $ & $ 3059 $ & American Sign Language (ASL)\\ 
HamNoSys~\citep{HamNoSys} & $ 4588 $ & $574$ & $573$ & Sign Transcriptive System\\
RWTH-PHOENIX-Weather~\citep{phoenix} & $ 2667 $ & $334$ & $333$ & Germany Sign Language (DGS)\\
\bottomrule
\end{tabular}}
\end{table}

\lstset{
    language=JSON,
    basicstyle=\ttfamily\footnotesize,
    numbers=left, 
    numberstyle=\tiny\color{gray}, 
    stepnumber=1,
    showstringspaces=false,
    breaklines=true,
    frame=lines, 
    backgroundcolor=\color{gray!8}, 
}
\begin{figure*}[b]
\centering
\begin{minipage}{\linewidth}
\begin{lstlisting}[caption=Templates used for pretraining and instruction-tuning., label=lst:merged_templates, basicstyle=\ttfamily\scriptsize]
{
  // Pretraining Template
  "Motion-to-Text": {
    "m2t": {
      "class": "m2t",
      "input": [
        "<Motion_Placeholder>"
      ],
      "output": [
        "<Caption_Placeholder>"
      ]
    }
  }
}

{
  // Instruction-Tuning Template
  "Motion-to-Text": {
    "caption": {
      "class": "m2t",
      "input": [
"Translate the American Sign Language represented by <Motion_Placeholder> to English.",
"Decipher the ASL communication in <Motion_Placeholder> and write it in English.",
"Rephrase the American Sign Language in <Motion_Placeholder> as spoken English.",
"Explain the meaning of the American Sign Language sequence <Motion_Placeholder> in English.",
...
      ],
      "output": [
        "<Caption_Placeholder>"
      ]
    }
  }
}
\end{lstlisting}
\end{minipage}
\vspace{-10pt}
\end{figure*}

\subsection{Dataset Detail}
\label{app:training_hyperparameters}
The details of the SignAvatar dataset are listed in List~\ref{lst:merged_templates}.

\section{More Qualitative Results}
\begin{figure*}[t]
\centering
\includegraphics[width=\linewidth]{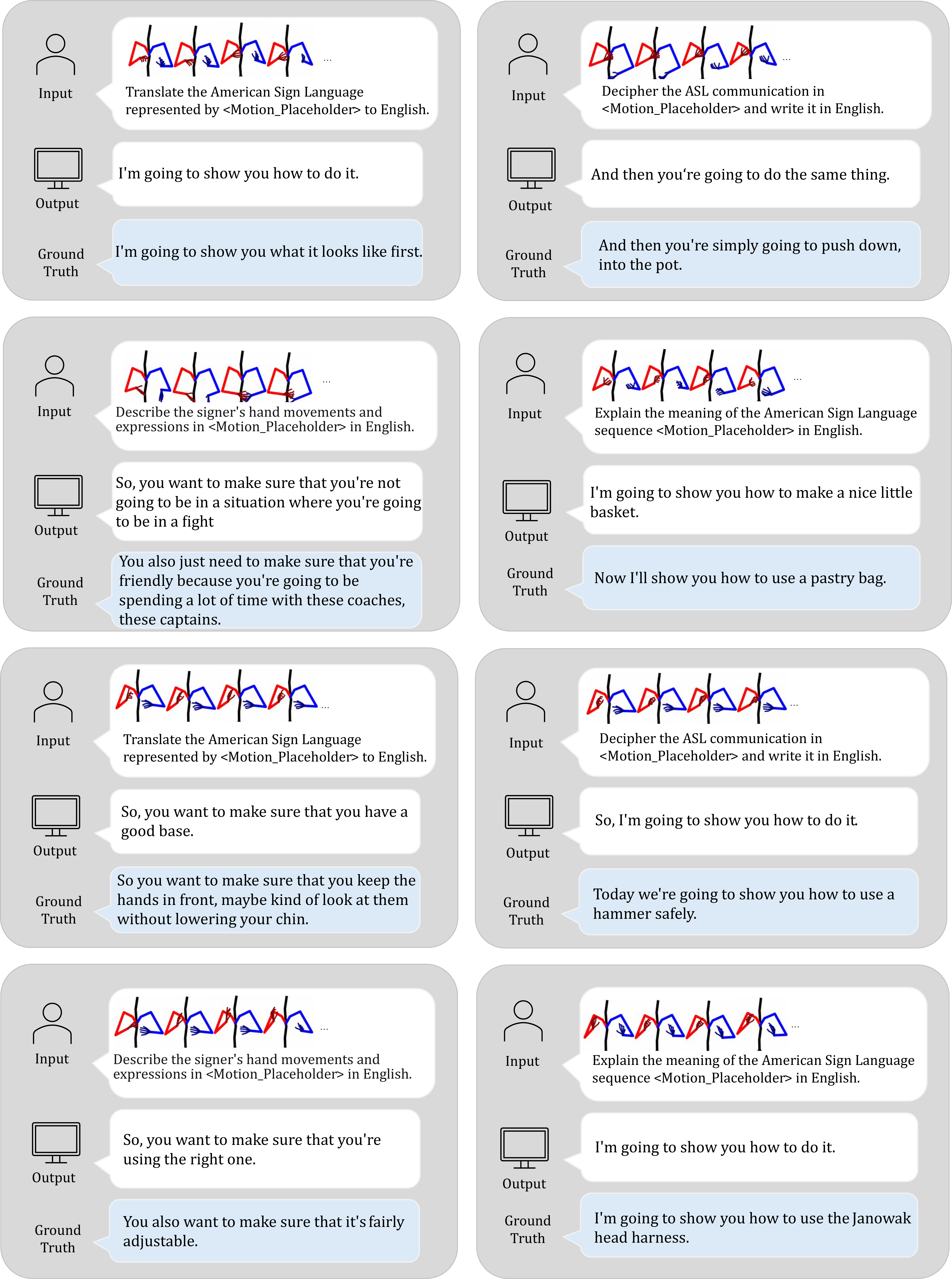}
\caption{Instruction-guided Sign Language Recognition (SLR) examples from our LSLM framework with Qwen backbone. Each case shows the model's translation given a gesture sequence and prompt, alongside the ground truth. Outputs may slightly differ in wording but preserve the core meaning.}

\label{fig:vis_slr}
\end{figure*}

We show more results from our LSLM with Qwen backbone in Fig~\ref{fig:vis_slr}.

%% file: main.bib
@String(CVPR= {IEEE Conf. Comput. Vis. Pattern Recog.})

@String(ICCV= {Int. Conf. Comput. Vis.})

@String(ECCV= {Eur. Conf. Comput. Vis.})

@String(AAAI = {AAAI})

@String(CVPR  = {CVPR})

@String(ICCV  = {ICCV})

@String(ECCV  = {ECCV})

@inproceedings{yu2024signavatars,
  title={SignAvatars: A large-scale 3D sign language holistic motion dataset and benchmark},
  author={Yu, Zhengdi and Huang, Shaoli and Cheng, Yongkang and Birdal, Tolga},
  booktitle={Proceedings of the European Conference on Computer Vision (ECCV)},
  pages={1--19},
  year={2024}
}

@inproceedings{gong2024llms,
  title={Llms are good sign language translators},
  author={Gong, Jia and Foo, Lin Geng and He, Yixuan and Rahmani, Hossein and Liu, Jun},
  booktitle={Proceedings of the IEEE/CVF Conference on Computer Vision and Pattern Recognition},
  pages={18362--18372},
  year={2024}
}

@INPROCEEDINGS{t2m,
  author={Guo, Chuan and Zou, Shihao and Zuo, Xinxin and Wang, Sen and Ji, Wei and Li, Xingyu and Cheng, Li},
  booktitle={2022 IEEE/CVF Conference on Computer Vision and Pattern Recognition (CVPR)}, 
  title={Generating Diverse and Natural 3D Human Motions from Text}, 
  year={2022},
  volume={},
  number={},
  pages={5142-5151},
  doi={10.1109/CVPR52688.2022.00509}}

@article{jiang2024motiongpt,
  title={Motiongpt: Human motion as a foreign language},
  author={Jiang, Biao and Chen, Xin and Liu, Wen and Yu, Jingyi and Yu, Gang and Chen, Tao},
  journal={Advances in Neural Information Processing Systems},
  volume={36},
  year={2024}
}

@article{wang2024motiongpt,
  title={MotionGPT-2: A General-Purpose Motion-Language Model for Motion Generation and Understanding},
  author={Wang, Yuan and Huang, Di and Zhang, Yaqi and Ouyang, Wanli and Jiao, Jile and Feng, Xuetao and Zhou, Yan and Wan, Pengfei and Tang, Shixiang and Xu, Dan},
  journal={arXiv preprint arXiv:2410.21747},
  year={2024}
}

@inproceedings{adamw,
  title={Decoupled Weight Decay Regularization},
  author={Ilya Loshchilov and Frank Hutter},
  booktitle={International Conference on Learning Representations},
  year={2017},
  url={https://api.semanticscholar.org/CorpusID:53592270}
}

@misc{MetaLlama3.2Blog2024,
  author       = {{Meta AI}},
  title        = {Llama 3.2: Revolutionizing edge AI and vision with open, customizable models},
  year         = {2024},
  month        = {September},
  day          = {25},
  howpublished = {Blog Post},
  url          = {https://ai.meta.com/blog/llama-3-2-connect-2024-vision-edge-mobile-devices/},
  note         = {Accessed: 2025-05-12}
}

@misc{qwen25technicalreport,
      title={Qwen2.5 Technical Report}, 
      author={Qwen and : and An Yang and Baosong Yang and Beichen Zhang and Binyuan Hui and Bo Zheng and Bowen Yu and Chengyuan Li and Dayiheng Liu and Fei Huang and Haoran Wei and Huan Lin and Jian Yang and Jianhong Tu and Jianwei Zhang and Jianxin Yang and Jiaxi Yang and Jingren Zhou and Junyang Lin and Kai Dang and Keming Lu and Keqin Bao and Kexin Yang and Le Yu and Mei Li and Mingfeng Xue and Pei Zhang and Qin Zhu and Rui Men and Runji Lin and Tianhao Li and Tianyi Tang and Tingyu Xia and Xingzhang Ren and Xuancheng Ren and Yang Fan and Yang Su and Yichang Zhang and Yu Wan and Yuqiong Liu and Zeyu Cui and Zhenru Zhang and Zihan Qiu},
      year={2025},
      eprint={2412.15115},
      archivePrefix={arXiv},
      primaryClass={cs.CL},
      url={https://arxiv.org/abs/2412.15115}, 
}

@Article{jimaging9110235,
AUTHOR = {Adão, Telmo and Oliveira, João and Shahrabadi, Somayeh and Jesus, Hugo and Fernandes, Marco and Costa, Angelo and Ferreira, Vania and Gonçalves, Martinho Fradeira and Lopéz, Miguel A. Guevara and Peres, Emanuel and Magalhaes, Luís Gonzaga},
TITLE = {Empowering Deaf-Hearing Communication: Exploring Synergies between Predictive and Generative AI-Based Strategies towards (Portuguese) Sign Language Interpretation},
JOURNAL = {Journal of Imaging},
VOLUME = {9},
YEAR = {2023},
NUMBER = {11},
ARTICLE-NUMBER = {235},
URL = {https://www.mdpi.com/2313-433X/9/11/235},
PubMedID = {37998082},
ISSN = {2313-433X},
DOI = {10.3390/jimaging9110235}
}

@article{kim2024leveraging,
  title={Leveraging the Power of MLLMs for Gloss-Free Sign Language Translation},
  author={Kim, Jungeun and Jeon, Hyeongwoo and Bae, Jongseong and Kim, Ha Young},
  journal={arXiv preprint arXiv:2411.16789},
  year={2024}
}

@inproceedings{camgoz2020sign,
  title={Sign language transformers: Joint end-to-end sign language recognition and translation},
  author={Camgoz, Necati Cihan and Koller, Oscar and Hadfield, Simon and Bowden, Richard},
  booktitle={Proceedings of the IEEE/CVF conference on computer vision and pattern recognition},
  pages={10023--10033},
  year={2020}
}

@inproceedings{bragg2019sign,
  title={Sign language recognition, generation, and translation: An interdisciplinary perspective},
  author={Bragg, Danielle and Koller, Oscar and Bellard, Mary and Berke, Larwan and Boudreault, Patrick and Braffort, Annelies and Caselli, Naomi and Huenerfauth, Matt and Kacorri, Hernisa and Verhoef, Tessa and others},
  booktitle={Proceedings of the 21st international ACM SIGACCESS conference on computers and accessibility},
  pages={16--31},
  year={2019}
}

@inproceedings{cui2017recurrent,
  title={Recurrent convolutional neural networks for continuous sign language recognition by staged optimization},
  author={Cui, Runpeng and Liu, Hu and Zhang, Changshui},
  booktitle={Proceedings of the IEEE conference on computer vision and pattern recognition},
  pages={7361--7369},
  year={2017}
}

@article{koller2020quantitative,
  title={Quantitative survey of the state of the art in sign language recognition},
  author={Koller, Oscar},
  journal={arXiv preprint arXiv:2008.09918},
  year={2020}
}

@article{cui2023spatial,
  title={Spatial--temporal transformer for end-to-end sign language recognition},
  author={Cui, Zhenchao and Zhang, Wenbo and Li, Zhaoxin and Wang, Zhaoqi},
  journal={Complex \& Intelligent Systems},
  volume={9},
  number={4},
  pages={4645--4656},
  year={2023},
  publisher={Springer}
}

@inproceedings{papineni-etal-2002-bleu,
    title = "{B}leu: a Method for Automatic Evaluation of Machine Translation",
    author = "Papineni, Kishore  and
      Roukos, Salim  and
      Ward, Todd  and
      Zhu, Wei-Jing",
    editor = "Isabelle, Pierre  and
      Charniak, Eugene  and
      Lin, Dekang",
    booktitle = "Proceedings of the 40th Annual Meeting of the Association for Computational Linguistics",
    month = jul,
    year = "2002",
    address = "Philadelphia, Pennsylvania, USA",
    publisher = "Association for Computational Linguistics",
    url = "https://aclanthology.org/P02-1040/",
    doi = "10.3115/1073083.1073135",
    pages = "311--318"
}

@inproceedings{lin-2004-rouge,
    title = "{ROUGE}: A Package for Automatic Evaluation of Summaries",
    author = "Lin, Chin-Yew",
    booktitle = "Text Summarization Branches Out",
    month = jul,
    year = "2004",
    address = "Barcelona, Spain",
    publisher = "Association for Computational Linguistics",
    url = "https://aclanthology.org/W04-1013/",
    pages = "74--81"
}

@inproceedings{Vedantam2015CIDEr,
  author    = {Vedantam, Ramakrishna and Lawrence Zitnick, C. and Parikh, Devi},
  title     = {{CIDEr: Consensus-based Image Description Evaluation}},
  booktitle = {Proceedings of the {IEEE} Conference on Computer Vision and Pattern Recognition ({CVPR})},
  pages     = {4566--4575},
  year      = {2015},
  month     = {June},
  address   = {Boston, MA, USA}
}

@conference{AMASS:ICCV:2019,
  title = {{AMASS}: Archive of Motion Capture as Surface Shapes},
  author = {Mahmood, Naureen and Ghorbani, Nima and Troje, Nikolaus F. and Pons-Moll, Gerard and Black, Michael J.},
  booktitle = {International Conference on Computer Vision},
  pages = {5442--5451},
  month = oct,
  year = {2019},
  month_numeric = {10}
}

@article{Plappert2016kit,
    author = {Matthias Plappert and Christian Mandery and Tamim Asfour},
    title = {The {KIT} Motion-Language Dataset},
    journal = {Big Data},
    publisher = {Mary Ann Liebert Inc},
    year = 2016,
    month = {dec},
    volume = {4},
    number = {4},
    pages = {236--252},
    url = {http://dx.doi.org/10.1089/big.2016.0028},
    doi = {10.1089/big.2016.0028},
}

@article{Davis2019HearingLoss,
  author    = {Davis, Adrian C. and Hoffman, Howard J.},
  title     = {{Hearing loss: rising prevalence and impact}},
  journal   = {Bulletin of the World Health Organization},
  year      = {2019},
  volume    = {97},
  number    = {10},
  pages     = {646--646A},
  doi       = {10.2471/BLT.19.224683},
  pmcid     = {PMC6796666}
}

@inproceedings{SMPL-X:2019,
  title = {Expressive Body Capture: {3D} Hands, Face, and Body from a Single Image},
  author = {Pavlakos, Georgios and Choutas, Vasileios and Ghorbani, Nima and Bolkart, Timo and Osman, Ahmed A. A. and Tzionas, Dimitrios and Black, Michael J.},
  booktitle = {Proceedings IEEE Conf. on Computer Vision and Pattern Recognition (CVPR)},
  pages     = {10975--10985},
  year = {2019}
}

@inproceedings{guo2022tm2t,
  title={Tm2t: Stochastic and tokenized modeling for the reciprocal generation of 3d human motions and texts},
  author={Guo, Chuan and Zuo, Xinxin and Wang, Sen and Cheng, Li},
  booktitle={European Conference on Computer Vision},
  pages={580--597},
  year={2022},
  organization={Springer}
}

@article{van2017neural,
  title={Neural discrete representation learning},
  author={Van Den Oord, Aaron and Vinyals, Oriol and others},
  journal={Advances in neural information processing systems},
  volume={30},
  year={2017}
}

@article{cai2023smpler,
  title={Smpler-x: Scaling up expressive human pose and shape estimation},
  author={Cai, Zhongang and Yin, Wanqi and Zeng, Ailing and Wei, Chen and Sun, Qingping and Yanjun, Wang and Pang, Hui En and Mei, Haiyi and Zhang, Mingyuan and Zhang, Lei and others},
  journal={Advances in Neural Information Processing Systems},
  volume={36},
  pages={11454--11468},
  year={2023}
}

@inproceedings{how2sign,
    title={{How2Sign: A Large-scale Multimodal Dataset for Continuous American Sign Language}},
    author={Duarte, Amanda and Palaskar, Shruti and Ventura, Lucas and Ghadiyaram, Deepti and DeHaan, Kenneth and
                   Metze, Florian and Torres, Jordi and Giro-i-Nieto, Xavier},
    booktitle={Conference on Computer Vision and Pattern Recognition (CVPR)},
    year={2021}
}

@misc{HamNoSys,
  title        = {{HamNoSys (Version 1997)}},
  author       = {{Institut für Deutsche Gebärdensprache und Kommunikation Gehörloser}},
  howpublished = {Webpage},
  organization = {Universität Hamburg, Institut für Deutsche Gebärdensprache und Kommunikation Gehörloser},
  year         = {1997},
}

@INPROCEEDINGS{phoenix,
  author={Camgoz, Necati Cihan and Hadfield, Simon and Koller, Oscar and Ney, Hermann and Bowden, Richard},
  booktitle={2018 IEEE/CVF Conference on Computer Vision and Pattern Recognition}, 
  title={Neural Sign Language Translation}, 
  year={2018},
  volume={},
  number={},
  pages={7784-7793},
  doi={10.1109/CVPR.2018.00812}}

@article{Zhang2024DeepLearningSLR,
  author    = {Zhang, Yanchi and Jiang, Xianwei},
  title     = {Recent Advances on Deep Learning for Sign Language Recognition},
  journal   = {Computer Modeling in Engineering \& Sciences},
  year      = {2024},
  volume    = {139},
  number    = {3},
  pages     = {2399--2450},
  doi       = {10.32604/cmes.2023.045731}
}

@inproceedings{Dinh2025SLRDataset,
  author    = {Dinh, Son Nguyen and Nguyen, Tuan Dung and Tran, Tri and Pham, Nguyen Dang Huy and Tran, Thuan Hieu and Tong, Ngoc Anh and Huy, Hoang Quang and Nguyen, Phi Le},
  title     = {{Sign Language Recognition: A Large-Scale Multi-View Dataset and Comprehensive Evaluation}},
  booktitle = {Proceedings of the IEEE/CVF Winter Conference on Applications of Computer Vision (WACV)},
  year      = {2025},
  month     = {February},
  note      = {Published online 28 Feb 2025 by OpenReview, official proceedings details may vary slightly.}
}

@inproceedings{Zheng2023CVTSLR,
  author    = {Zheng, Jiangbin and Wang, Yile and Tan, Cheng and Li, Siyuan and Wang, Ge and Xia, Jun and Chen, Yidong and Li, Stan Z.},
  title     = {{CVT-SLR: Contrastive Visual-Textual Transformation for Sign Language Recognition with Variational Alignment}},
  booktitle = {Proceedings of the IEEE/CVF Conference on Computer Vision and Pattern Recognition (CVPR)},
  year      = {2023},
  pages     = {14977-14987}
}

@article{Kumari2024HybridSLR,
  author    = {Kumari, D. and Anand, R. S.},
  title     = {{Isolated Video-Based Sign Language Recognition Using a Hybrid CNN-LSTM Framework Based on Attention Mechanism}},
  journal   = {Electronics},
  year      = {2024},
  volume    = {13},
  number    = {7},
  pages     = {1229},
  doi       = {10.3390/electronics13071229}
}

@article{Huang2024ResNetLSTM,
  author    = {Huang, Jiayu and Chouvatut, Varin},
  title     = {{Video-Based Sign Language Recognition via ResNet and LSTM Network}},
  journal   = {Journal of Imaging},
  year      = {2024},
  volume    = {10},
  number    = {6},
  pages     = {149},
  doi       = {10.3390/jimaging10060149}
}

@inproceedings{Zuo2024OnlineCSLRT,
  author    = {Zuo, Ronglai and Liu, Zekang and Hu, Lianyu and Zheng, Jiangbin and Gao, Liqing and Wei, Fangyun and Mak, Brian},
  title     = {{Towards Online Continuous Sign Language Recognition and Translation}},
  booktitle = {Proceedings of the 2024 Conference on Empirical Methods in Natural Language Processing (EMNLP)},
  year      = {2024},
  publisher = {Association for Computational Linguistics},
  month     = {November},
  note      = {Page numbers may vary with final publication; check ACL Anthology for updates.}
}

@misc{Ranjbar2024IntraInterGloss,
  author    = {Ranjbar, Hamid and Taheri, Ahmad},
  title     = {{Continuous Sign Language Recognition Using Intra-inter Gloss Attention}},
  year      = {2024},
  eprint    = {2406.18333},
  archivePrefix = {arXiv},
  primaryClass = {cs.CV},
  month     = {June}
}

@inproceedings{Hu2023SelfEmphasizingCSLR,
  author    = {Hu, Lianyu and Gao, Liqing and Liu, Zekang and Feng, Wei},
  title     = {{Self-Emphasizing Network for Continuous Sign Language Recognition}},
  booktitle = {Proceedings of the AAAI Conference on Artificial Intelligence},
  year      = {2023},
  volume    = {37},
  number    = {1},
  pages     = {1018--1026},
  doi       = {10.1609/aaai.v37i1.25199}
}

@article{chen2025snapmoji,
  title={Snapmoji: Instant Generation of Animatable Dual-Stylized Avatars},
  author={Chen, Eric M and Liu, Di and Ma, Sizhuo and Vasilkovsky, Michael and Zhou, Bing and Gao, Qiang and Wang, Wenzhou and Luo, Jiahao and Metaxas, Dimitris N and Sitzmann, Vincent and others},
  journal={arXiv preprint arXiv:2503.11978},
  year={2025}
}

@article{he2024dice,
  title={Dice: Discrete inversion enabling controllable editing for multinomial diffusion and masked generative models},
  author={He, Xiaoxiao and Han, Ligong and Dao, Quan and Wen, Song and Bai, Minhao and Liu, Di and Zhang, Han and Min, Martin Renqiang and Juefei-Xu, Felix and Tan, Chaowei and others},
  journal={arXiv preprint arXiv:2410.08207},
  year={2024}
}

@InProceedings{Luo_2025_ICCV,
    author    = {Luo, Jiahao and Wang, Chaoyang and Vasilkovsky, Michael and Shakhrai, Vladislav and Liu, Di and Zhuang, Peiye and Tulyakov, Sergey and Wonka, Peter and Lee, Hsin-Ying and Davis, James and Wang, Jian},
    title     = {T2Bs: Text-to-Character Blendshapes via Video Generation},
    booktitle = {Proceedings of the IEEE/CVF International Conference on Computer Vision (ICCV)},
    month     = {October},
    year      = {2025},
    pages     = {13625-13637}
}

@inproceedings{xia2025visiar,
  title={VISIAR: Empower MLLM for visual story ideation},
  author={Xia, Zhaoyang and Sarkhel, Somdeb and Tanjim, Mehrab and Petrangeli, Stefano and Dasgupta, Ishita and Chen, Yuxiao and Xu, Jinxuan and Liu, Di and Mitra, Saayan and Metaxas, Dimitris N},
  booktitle={Findings of the Association for Computational Linguistics: ACL 2025},
  pages={18384--18402},
  year={2025}
}

@inproceedings{liu2025lucas,
  title={LUCAS: Layered Universal Codec Avatars},
  author={Liu, Di and Deng, Teng and Nam, Giljoo and Rong, Yu and Pidhorskyi, Stanislav and Li, Junxuan and Saragih, Jason and Metaxas, Dimitris N and Cao, Chen},
  booktitle={Proceedings of the Computer Vision and Pattern Recognition Conference},
  pages={21127--21137},
  year={2025}
}

@inproceedings{gao2025show,
  title={Show and Segment: Universal Medical Image Segmentation via In-Context Learning},
  author={Gao, Yunhe and Liu, Di and Li, Zhuowei and Li, Yunsheng and Chen, Dongdong and Zhou, Mu and Metaxas, Dimitris N},
  booktitle={Proceedings of the Computer Vision and Pattern Recognition Conference},
  pages={20830--20840},
  year={2025}
}

@article{li2025hidden,
  title={The hidden life of tokens: Reducing hallucination of large vision-language models via visual information steering},
  author={Li, Zhuowei and Shi, Haizhou and Gao, Yunhe and Liu, Di and Wang, Zhenting and Chen, Yuxiao and Liu, Ting and Zhao, Long and Wang, Hao and Metaxas, Dimitris N},
  journal={arXiv preprint arXiv:2502.03628},
  year={2025}
}

@article{dao2025improved,
  title={Improved training technique for latent consistency models},
  author={Dao, Quan and Doan, Khanh and Liu, Di and Le, Trung and Metaxas, Dimitris},
  journal={arXiv preprint arXiv:2502.01441},
  year={2025}
}

@article{li2024implicit,
  title={Implicit in-context learning},
  author={Li, Zhuowei and Xu, Zihao and Han, Ligong and Gao, Yunhe and Wen, Song and Liu, Di and Wang, Hao and Metaxas, Dimitris N},
  journal={arXiv preprint arXiv:2405.14660},
  year={2024}
}

@article{zhangli2024resolving,
  title={Resolving Inconsistent Semantics in Multi-Dataset Image Segmentation},
  author={Zhangli, Qilong and Liu, Di and Aich, Abhishek and Metaxas, Dimitris and Schulter, Samuel},
  journal={arXiv preprint arXiv:2409.09893},
  year={2024}
}

@inproceedings{zhangli2024layout,
  title={Layout-agnostic scene text image synthesis with diffusion models},
  author={Zhangli, Qilong and Jiang, Jindong and Liu, Di and Yu, Licheng and Dai, Xiaoliang and Ramchandani, Ankit and Pang, Guan and Metaxas, Dimitris N and Krishnan, Praveen},
  booktitle={2024 IEEE/CVF Conference on Computer Vision and Pattern Recognition (CVPR)},
  pages={7496--7506},
  year={2024},
  organization={IEEE Computer Society}
}

@inproceedings{liu2024instantaneous,
  title={Instantaneous perception of moving objects in 3d},
  author={Liu, Di and Zhuang, Bingbing and Metaxas, Dimitris N and Chandraker, Manmohan},
  booktitle={Proceedings of the IEEE/CVF conference on computer vision and pattern recognition},
  pages={19573--19583},
  year={2024}
}

@inproceedings{gao2024training,
  title={Training like a medical resident: Context-prior learning toward universal medical image segmentation},
  author={Gao, Yunhe},
  booktitle={Proceedings of the IEEE/CVF Conference on Computer Vision and Pattern Recognition},
  pages={11194--11204},
  year={2024}
}

@inproceedings{han2024proxedit,
  title={Proxedit: Improving tuning-free real image editing with proximal guidance},
  author={Han, Ligong and Wen, Song and Chen, Qi and Zhang, Zhixing and Song, Kunpeng and Ren, Mengwei and Gao, Ruijiang and Stathopoulos, Anastasis and He, Xiaoxiao and Chen, Yuxiao and others},
  booktitle={Proceedings of the IEEE/CVF Winter Conference on Applications of Computer Vision},
  pages={4291--4301},
  year={2024}
}

@inproceedings{wen2024second,
  title={Second-order graph odes for multi-agent trajectory forecasting},
  author={Wen, Song and Wang, Hao and Liu, Di and Zhangli, Qilong and Metaxas, Dimitris},
  booktitle={Proceedings of the IEEE/CVF Winter Conference on Applications of Computer Vision},
  pages={5101--5110},
  year={2024}
}

@inproceedings{li2024steering,
  title={Steering prototypes with prompt-tuning for rehearsal-free continual learning},
  author={Li, Zhuowei and Zhao, Long and Zhang, Zizhao and Zhang, Han and Liu, Di and Liu, Ting and Metaxas, Dimitris N},
  booktitle={Proceedings of the IEEE/CVF Winter Conference on Applications of Computer Vision},
  pages={2523--2533},
  year={2024}
}

@article{liu2023lepard,
  title={Lepard: Learning explicit part discovery for 3d articulated shape reconstruction},
  author={Liu, Di and Stathopoulos, Anastasis and Zhangli, Qilong and Gao, Yunhe and Metaxas, Dimitris},
  journal={Advances in Neural Information Processing Systems},
  volume={36},
  pages={54187--54198},
  year={2023}
}

@inproceedings{liu2023deformer,
  title={Deformer: Integrating transformers with deformable models for 3d shape abstraction from a single image},
  author={Liu, Di and Yu, Xiang and Ye, Meng and Zhangli, Qilong and Li, Zhuowei and Zhang, Zhixing and Metaxas, Dimitris N},
  booktitle={Proceedings of the IEEE/CVF International Conference on Computer Vision},
  pages={14236--14246},
  year={2023}
}

@article{liu2023deep,
  title={Deep deformable models: Learning 3d shape abstractions with part consistency},
  author={Liu, Di and Zhao, Long and Zhangli, Qilong and Gao, Yunhe and Liu, Ting and Metaxas, Dimitris N},
  journal={arXiv preprint arXiv:2309.01035},
  year={2023}
}

@article{martin2023deep,
  title={Deep learning segmentation of the right ventricle in cardiac MRI: the M\&Ms challenge},
  author={Mart{\'\i}n-Isla, Carlos and Campello, V{\'\i}ctor M and Izquierdo, Cristian and Kushibar, Kaisar and Sendra-Balcells, Carla and Gkontra, Polyxeni and Sojoudi, Alireza and Fulton, Mitchell J and Arega, Tewodros Weldebirhan and Punithakumar, Kumaradevan and others},
  journal={IEEE Journal of Biomedical and Health Informatics},
  volume={27},
  number={7},
  pages={3302--3313},
  year={2023},
  publisher={IEEE}
}

@inproceedings{he2023dealing,
  title={Dealing with heterogeneous 3d mr knee images: A federated few-shot learning method with dual knowledge distillation},
  author={He, Xiaoxiao and Tan, Chaowei and Liu, Bo and Si, Liping and Yao, Weiwu and Zhao, Liang and Liu, Di and Zhangli, Qilong and Chang, Qi and Li, Kang and others},
  booktitle={2023 IEEE 20th International Symposium on Biomedical Imaging (ISBI)},
  pages={1--5},
  year={2023},
  organization={IEEE}
}

@inproceedings{liu2022transfusion,
  title={Transfusion: multi-view divergent fusion for medical image segmentation with transformers},
  author={Liu, Di and Gao, Yunhe and Zhangli, Qilong and Han, Ligong and He, Xiaoxiao and Xia, Zhaoyang and Wen, Song and Chang, Qi and Yan, Zhennan and Zhou, Mu and others},
  booktitle={International conference on medical image computing and computer-assisted intervention},
  pages={485--495},
  year={2022},
  organization={Springer}
}

@inproceedings{chang2022deeprecon,
  title={Deeprecon: Joint 2d cardiac segmentation and 3d volume reconstruction via a structure-specific generative method},
  author={Chang, Qi and Yan, Zhennan and Zhou, Mu and Liu, Di and Sawalha, Khalid and Ye, Meng and Zhangli, Qilong and Kanski, Mikael and Al’Aref, Subhi and Axel, Leon and others},
  booktitle={International Conference on Medical Image Computing and Computer-Assisted Intervention},
  pages={567--577},
  year={2022},
  organization={Springer}
}

@inproceedings{zhangli2022region,
  title={Region proposal rectification towards robust instance segmentation of biological images},
  author={Zhangli, Qilong and Yi, Jingru and Liu, Di and He, Xiaoxiao and Xia, Zhaoyang and Chang, Qi and Han, Ligong and Gao, Yunhe and Wen, Song and Tang, Haiming and others},
  booktitle={International Conference on Medical Image Computing and Computer-Assisted Intervention},
  pages={129--139},
  year={2022},
  organization={Springer}
}

@inproceedings{liu2021refined,
  title={Refined deep layer aggregation for multi-disease, multi-view \& multi-center cardiac mr segmentation},
  author={Liu, Di and Yan, Zhennan and Chang, Qi and Axel, Leon and Metaxas, Dimitris N},
  booktitle={International Workshop on Statistical Atlases and Computational Models of the Heart},
  pages={315--322},
  year={2021},
  organization={Springer}
}

@inproceedings{liu2021label,
  title={Label super resolution for 3d magnetic resonance images using deformable u-net},
  author={Liu, Di and Liu, Jiang and Liu, Yihao and Tao, Ran and Prince, Jerry L and Carass, Aaron},
  booktitle={Medical Imaging 2021: Image Processing},
  volume={11596},
  pages={606--611},
  year={2021},
  organization={SPIE}
}
